# Synthetic Data Generation by Supervised Neural Gas Network for Physiological Emotion Recognition Data

S. Muhammad Hossein Mousavi
Tehran, Iran
ORCID: 0000-0001-6906-2152
s.muhammad.hossein.mousavi@gmail.com

**Abstract**

Data scarcity remains a significant challenge in the field of emotion recognition using physiological signals, as acquiring comprehensive and diverse datasets is often prevented by privacy concerns and logistical constraints. This limitation restricts the development and generalization of robust emotion recognition models, making the need for effective synthetic data generation methods more critical. Emotion recognition from physiological signals such as EEG, ECG, and GSR plays a pivotal role in enhancing human-computer interaction and understanding human affective states. Utilizing these signals, this study introduces an innovative approach to synthetic data generation using a Supervised Neural Gas (SNG) network, which has demonstrated noteworthy speed advantages over established models like Conditional VAE, Conditional GAN, diffusion model, and Variational LSTM. The Neural Gas network, known for its adaptability in organizing data based on topological and feature-space proximity, provides a robust framework for generating real-world-like synthetic datasets that preserve the intrinsic patterns of physiological emotion data. Our implementation of the SNG efficiently processes the input data, creating synthetic instances that closely mimic the original data distributions, as demonstrated through comparative accuracy assessments. In experiments, while our approach did not universally outperform all models, it achieved superior performance against most of the evaluated models and offered significant improvements in processing time. These outcomes underscore the potential of using SNG networks for fast, efficient, and effective synthetic data generation in emotion recognition applications.

## 1. Introduction

Emotion recognition [1, 54, 55] is the process of identifying human emotions using various data inputs and algorithms, playing a critical role in enhancing Human-Computer Interaction [2, 20]. This capability is crucial for advancing fields such as personalized marketing [18], mental health monitoring [19], and adaptive learning systems [3], where understanding human feelings can significantly optimize interactions and outcomes. Physiological signals, such as electroencephalograms (EEG), electrocardiograms (ECG), and galvanic skin response (GSR), are key in this domain due to their direct measurement of bodily states that reflect emotional conditions [3]. Figure 2 depicts EEG, ECG, and GSR sample signals belonging to the joy emotion state. These signals are particularly valuable in applications like lie detection, patient monitoring in healthcare, and enhancing user engagement in gaming and virtual reality, where accurate emotion detection can greatly enhance user experience and outcomes [3, 4]. The use of EEG, ECG, and GSR in emotion recognition taps into unique aspects of physiological responses, enabling the detection of nuanced emotional states with a level of precision not achievable through behavioral analysis alone. EEG measures electrical activity in the brain to reveal patterns associated with different emotional states, while ECG assesses heart rate variability as an indicator of emotional arousal. GSR monitors changes in skin conductance, which varies with emotional intensity [3, 4]. Together, these signals provide a comprehensive physiological footprint of emotional states. However, a significant challenge in utilizing these signals for emotion recognition is data scarcity [5-8]. The difficulties in collecting large, diverse, and representative datasets stem from privacy concerns, high collection costs, and the technical complexity of accurately capturing and processing these signals. This scarcity prevents the development of robust models that perform well across different populations and environments. Addressing this issue is crucial for the advancement of reliable and generalizable emotion recognition technologies, underscoring the need for innovative solutions like Synthetic Data Generation (SDG) [9, 6, 10] to bridge the data gap.

SDG involves creating artificial datasets that statistically mirror real-world data, offering a promising solution to the issue of data scarcity in emotion recognition from physiological signals. By employing algorithms capable of learning and replicating the complex patterns found in actual physiological data, synthetic data can be generated to enhance existing datasets without compromising individual privacy. This method is particularly beneficial in fields where data collection is limited by ethical concerns, such as in health-related research [11]. In the context of emotion recognition, synthetic datasets allow researchers and developers to train and test algorithms with a broader range of data inputs, increasing the robustness and accuracy of predictive models. Moreover, these synthetic datasets help overcome the barriers of limited sample sizes and lack of diversity in training data, thus supporting the development of emotion recognition systems that are both effective and adaptable across various real-world scenarios. Figure 1 illustrates the number of publications per year for emotion recognition, synthetic data generation, and physiological signal augmentation topics extracted from the PubMed dataset[1]. All of them show a growing number of publications in years.

The Neural Gas Network (NGN) [17] is a type of artificial neural network that adapts to input data without a predetermined network structure, efficiently organizing itself to reflect the topology of the data it processes. This flexibility makes NGN particularly useful in applications such as vector quantization [15], clustering, dimensionality reduction[2], image segmentation [12], and feature extraction [13]. The inherent adaptability of NGN to different data distributions allows it to capture complex patterns in high-dimensional spaces effectively. Extending NGN into a supervised learning framework enhances its applicability to tasks involving classification and prediction. In a supervised setting, NGN can utilize labeled data, guiding the network's adaptation process more precisely toward task-specific objectives. This makes Supervised Neural Gas (SNG) [15] well-suited for applications where precise categorization of complex patterns is crucial, such as in text categorization, image recognition, and bioinformatics [16].

In the synthetic data generation for emotion recognition using physiological signals like EEG, ECG, and GSR, SNG offers distinct advantages. By integrating the classification labels directly into the learning process, SNG can generate synthetic data that not only resembles the original data in terms of distribution but also aligns accurately with specific emotional states. This capability is critical for developing robust emotion recognition systems that require extensive, varied, and accurately labeled datasets for training. Compared to traditional methods, SNG provides a more direct mechanism for controlling the generation process based on the topology and distribution of input data, resulting in faster processing times and potentially higher accuracy in reflecting complex physiological and emotional correlations. This makes SNG an effective tool in overcoming the challenges of data scarcity and enhancing the performance of emotion recognition systems. According to our research, this is the first time the SNG has been used for SDG in this research. Here, we are looking forward to answering the following research questions. RQ1: Is SNG capable of generating real-world-like emotion recognition physiological EEG, ECG, and GSR signals by capturing complex relations in between signals? RQ2: Can SNG-generated data effectively outperform other SDG methods in terms of diversity, accuracy, and training speed? RQ3: Does SNG SDG address the challenge of data scarcity in emotion recognition using physiological signals such as EEG, ECG, GSR, and probably other physiological signals?

We have successfully applied Supervised Neural Gas (SNG) for synthetic data generation in emotion recognition, pioneering its use in creating realistic physiological datasets. Our approach has effectively bridged the gap in data scarcity, enhancing the robustness and training speed of emotion recognition systems. This achievement demonstrates the practical benefits and versatility of SNG, affirming its value as a powerful tool in effective computing research. The result section explains the achievement in detail.

---

[1] https://pubmed.ncbi.nlm.nih.gov/
[2] https://github.com/SeyedMuhammadHosseinMousavi/Neural-Gas-Network-Toolbox

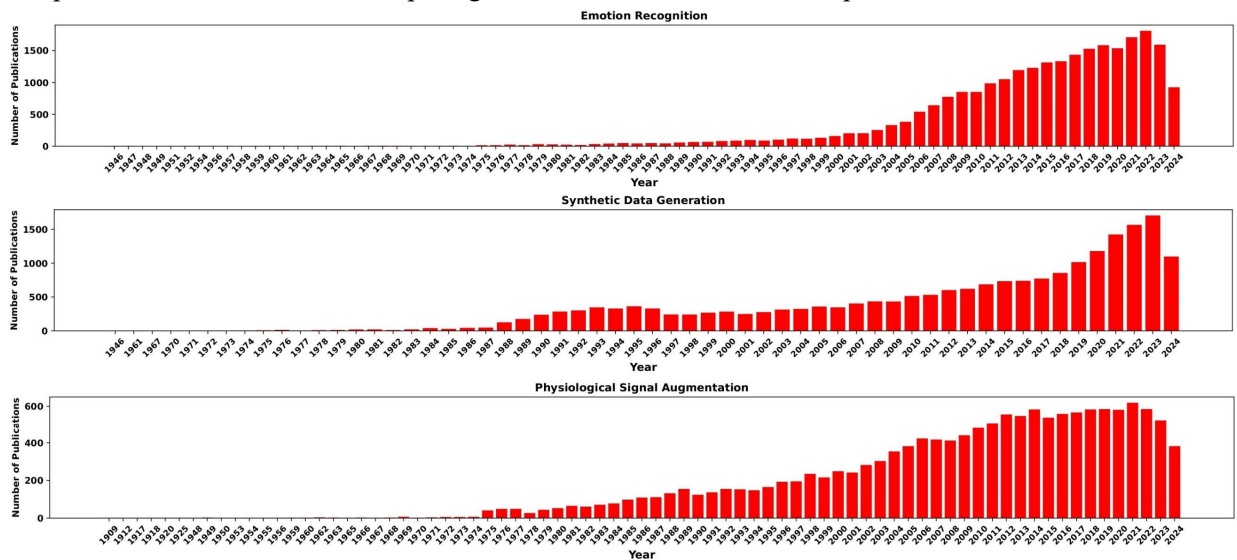

Figure 1. The number of publications per year for emotion recognition, synthetic data generation, and physiological signal augmentation topics extracted from the PubMed dataset

Section 2 pays to related works done by other researchers in the field of SDG of physiological signals in emotion recognition. Section 3 covers the theoretical background, section 4 covers our proposed method, and section 5 covers our evaluations and results. Finally, the conclusion packs up our contribution. You can find the GitHub repository of our contribution's implementation by Python in the footnote[3].

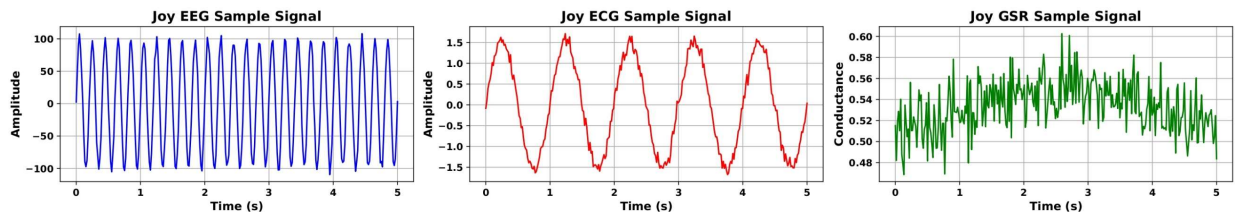

Figure 2. EEG, ECG, and GSR sample signals belonging to the joy emotional state

## 2. Related Works

This section covers research conducted by other researchers in the field of SDG of physiological signals, especially in the field of emotion recognition. To save space, the chronicle is reported in Table 1 in an organized manner.

Table 1. Prior related works regarding SDG of physiological signals in emotion recognition

| # | Author(s) | Subject | Challenge | Contribution/Solution | Year | Cite |
|---|-----------|---------|-----------|----------------------|------|------|
| 1 | Nita, Sihem, et al | ECG SDG | Data Scarcity | Using the CNN algorithm for emotional ECG signals SDG by the DREAMER database | 2022 | [21] |
| 2 | Guo, Gengyuan, et al | ECG, GSR, and respiration (RSP) SDG | Data Scarcity | Using CNN-SVM technique for SDG of ECG, GSR, and RSP signals for emotion recognition data | 2022 | [22] |
| 3 | Chen, Yu, Rui Chang | EEG SDG | Data Scarcity | Using SMOTE CNN algorithm for EEG signals SDG for emotion recognition data on DEAP dataset | 2021 | [23] |

---

[3] The GitHub code repository of this contribution is available at: https://github.com/SeyedMuhammadHosseinMousavi/Synthetic-Data-Generation-by-Supervised-Neural-Gas-Network

| 4 | Ari, Berna, et al | EEG SDG | Data Scarcity | Proposed Extreme Learning Machine Wavelet Auto Encoder (ELM-W-AE) for EEG signals SDG for emotion recognition data | 2022 | [24] |
|---|---|---|---|---|---|---|
| 5 | Nasrallah, Chawki, et al | Electromyography (EMG) SDG | Data Scarcity | Conditional GAN has been used for EMG signals SDG or augmentation for emotion recognition data | 2023 | [25] |
| 6 | Wang, Fang, et al | EEG SDG | Data Scarcity | Using CNN for EEG signals SDG and for emotion recognition data | 2018 | [26] |
| 7 | Hasnul, M.A., et al | ECG SDG | Data Scarcity | They used multiple filter techniques to augment ECG emotional signals and evaluate using different classifiers on three datasets of DREAMER, A2ES, and AMIGOS | 2023 | [27] |
| 8 | Kalashami, M.P., et al | EEG SDG | Data Scarcity | Using Conditional Wasserstein GAN (CWGAN) for EEG signal SDG for emotion recognition data on the DEAP dataset | 2022 | [28] |
| 9 | Furdui, Andrei, et al | GSR and ECG SDG | Data Scarcity | Using Auxiliary Conditioned Wasserstein Generative Adversarial Network with Gradient Penalty (AC-WGAN-GP) to synthesize/augment GSR and ECG signals for emotion recognition data | 2021 | [29] |
| 10 | Grossi, A., et al | Photoplethysmogram (PPG) SDG | Data Scarcity | Using Complete Ensemble Empirical Mode Decomposition with Adaptive Noise (CEEMDAN) for PPG signals SDG | 2023 | [30] |
| 11 | Adib, Edmond., et al | ECG SDG | Data Scarcity | Using GAN for augmenting ECG signals | 2021 | [31] |
| 12 | Hazra, Debapriya., et al | ECG, EEG, EMG, PPG SDG | Data Scarcity | They propose a novel GAN model, named SynSigGAN, for automating the generation of any synthetic physiological signals | 2020 | [32] |
| 13 | Silva, Diogo., et al | Heart Rate (HR) SDG | Data Scarcity | They used a stochastic system of Gaussian copulas integrated in a Markov chain to augment HR signals | 2020 | [33] |
| 14 | Pereira, Diogo Filipe., et al | ECG and ballistocardiography (BCG) SDG | Data Scarcity | Using Gaussian Copula for ECG and BCG signals SDG | 2019 | [34] |
| 15 | Saldanha, Jane, et al | RSP SDG | Data Scarcity | Using Variational Autoencoders like Multilayer Perceptron VAE (MLP-VAE), Convolutional VAE (CVAE), and Conditional VAE for RSP SDG | 2022 | [35] |
| 16 | Soingern, Nutapol, et al | EEG SDG | Data Scarcity | Using the diffusion model method to augment EEG signals | 2023 | [36] |
| 17 | Siddhad, Gourav, et al | EEG SDG | Data Scarcity | Using the diffusion model method to augment EEG signals on DEAP dataset for emotion recognition data | 2024 | [37] |
| 18 | Takahashi, Kahoko, et al | EEG SDG | Data Scarcity | Using the LSTM algorithm to augment EEG signals | 2022 | [38] |
| 19 | Li, Xiaomin., et al | ECG SDG | Data Scarcity | They proposed TTS-CGAN, a Transformer-based Time-Series Conditional GAN to augment ECG signals from the PTB Diagnostic ECG dataset | 2022 | [39] |

## 3. Theoretical Background

The Neural Gas Network (NGN) [17] is a type of artificial neural network known for its adaptability and self-organizing capabilities. Unlike traditional neural networks, NGN does not require a pre-defined network structure. Instead, it organically arranges itself to mirror the topology of the input data it processes. This flexibility allows NGN to efficiently handle various applications such as vector quantization, clustering, dimensionality reduction, image segmentation, and feature extraction. NGN's ability to adapt to different data distributions enables it to effectively capture complex patterns in high-dimensional spaces, making it highly effective for tasks that involve intricate data structures.

To convert a Neural Gas Network (NGN) into a supervised learning framework, creating a Supervised Neural Gas (SNG) [15] involves integrating target labels directly into the learning process. The primary step is to modify the typical unsupervised training method of NGN, which focuses on finding the optimal representation of data without considering any external labels. In SNG, during each iteration of the training process, not only are the nearest neurons to the input data points activated but they are also associated with specific target labels from the training dataset. This association allows the network to adjust its weights not just based on the proximity of the data points but also based on the correctness of the label prediction. Both NGN and SNG use competitive learning, where all neurons in the network compete to be closer to the current input data point. The winning neuron (the one closest to the input) and its neighbors (defined by a neighborhood function) are adjusted to be even closer to that data point. Over time, this competitive process results in a network that reflects the topology and, in the case of SNG, the class structure of the input space. NGN and SNG generally consist of two layers. The input layer receives the input features and connects each feature to every neuron in the next layer. Also, the competition layer is where neurons compete to be closest to the input vector; each neuron adjusts based on its distance from the input and is associated with a class label. There are no hidden layers as found in more traditional neural networks. The second layer comprises neurons that compete to be closer to the input data through a process that adjusts their weights. The learning rate and the neighborhood function, crucial parameters in NGN, are customized to decrease over time in a way that reflects both the error in label prediction and the topological accuracy. Additionally, the cost function in SNG is designed to incorporate a penalty for misclassification, thus aligning the neuron adjustments more closely with the supervised learning objectives. This method effectively transforms the NGN into an SNG, enabling it to perform classification tasks by utilizing the structured adaptation of neurons in response to labeled data, enhancing both the accuracy and applicability of the network in complex pattern recognition scenarios. The typical learning update equation for a Neural Gas Network (NGN) is as follows:

$$NGN = w_i(t+1) = w_i(t) + \epsilon(t) \cdot h_\lambda(t, k(i,x)) \cdot (x - w_i(t)) \tag{1}$$

Where, $w_i(t)$ is the weight vector of the $i$-th neuron at the time $t$. $x$ is the current input vector. $\epsilon(t)$ is the learning rate at the time $t$, which decreases over time. $h_\lambda(t, k(i,x))$ is the neighborhood function around the winning neuron. This function decreases with increasing rank $k(i,x)$ of the neuron $i$ when ordered by distance from the input vector $x$. The function is also dependent on a parameter $\lambda(t)$, which is a measure of the neighborhood size that decreases over time. $k(i,x)$ is the rank of neuron $i$ in terms of its distance from the input vector $x$, with the closest neuron having the rank 0. Also, $t+1$ represents the next time step.

The adaptation of NGN to a Supervised Neural Gas (SNG) involves modifying the update rule to include the influence of the target label. The modified update equation is as follows:

$$SNG = w_i(t+1) = w_i(t) + \epsilon(t) \cdot h_\lambda(t, k(i,x)) \cdot (x - w_i(t)) + \alpha(t) \cdot g(t, y, y_i) \cdot (y - y_i) \tag{2}$$

Where, $y$ is the actual target label for the input $x$. $y_i$ is the predicted label from the neuron $i$. $\alpha(t)$ is an additional learning rate parameter governing the adaptation based on the label mismatch. Furthermore, $g(t, y, y_i)$ is a function that measures the error in label prediction, which could be a simple delta function $\delta(y, y_i)$ indicating 1 when $y \neq y_i$ and 0 otherwise. Figure 3 represents how NGN fills the topology of the destination. There are 200 blue dot samples as the destination topology and 150 red dot samples or neurons for NGN, in which over 300 iterations of NGN tris fill the shape with those neurons. That 150 neurons is exactly the desired number of synthetic samples that we are looking for. This figure shows the potential of NGN to generate similar-like samples by fitting the topology.

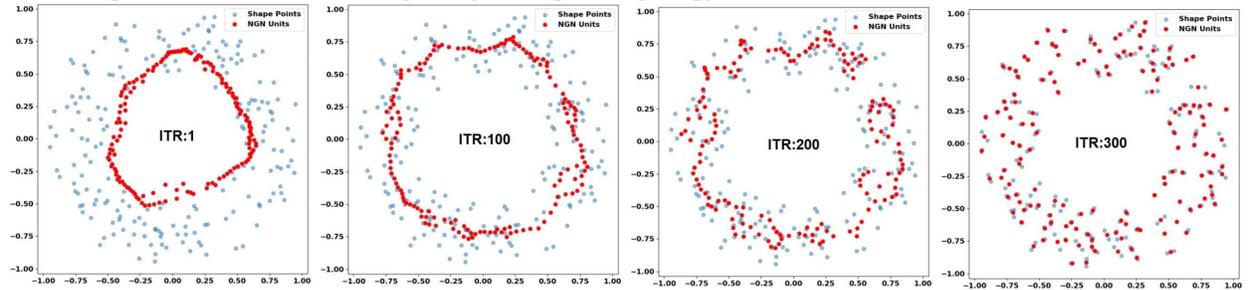

Figure 3. NGN topology fitting process over 300 iterations using 150 neurons

## 4. Proposed Method

Using SNG for SDG involves two stages of training and generation based on two main equations, (3) and (7). All steps are described below in detail.

$$SNG\ Train = w_{c,i}(t+1) = w_{c,i}(t) + \eta(t) \cdot h_\lambda(t, k(c, i, x)) \cdot (x - w_{c,i}(t)) \tag{3}$$

Where, $w_{c,i}(t)$ is the weight vector of the $i$-th neuron for class $c$ at iteration $t$. These weights represent the prototype vectors that are being adapted to the data. $x$ is the input vector from the training dataset associated with the class $c$. Also, $\eta(t)$ is the learning rate at iteration $t$, which decreases over time. It governs how much the neuron weights are updated in each step and is calculated as:

$$\eta(t) = \eta_{\text{start}} \left(\frac{\eta_{\text{end}}}{\eta_{\text{start}}}\right)^{\frac{t}{\max i \, \text{iter}}} \quad (4)$$

$h_\lambda(t, k(c, i, x))$ is the neighborhood function, which decreases with the rank $k(c, i, x)$. This function weakens the influence of the input vector based on the rank of the neuron within the class:

$$h_\lambda(t, k) = e^{-\frac{k}{\lambda(t)}} \quad (5)$$

$\lambda(t)$ is the neighborhood range, which also decreases over time, controls the extent of the local neighborhood around the best-matching unit that gets updated:

$$\lambda(t) = \lambda_{\text{start}} \left(\frac{\lambda_{\text{end}}}{\lambda_{\text{start}}}\right)^{\frac{t}{\max \text{iter}}} \quad (6)$$

Finally, for the generating synthetic sample, we have the following:

$$x_{\text{synthetic}} = w_{c,i} + \mathcal{N}(0, \sigma^2) \quad (7)$$

Where:

$w_{c,i}$ is the final learned neuron weight after the training phase, representing a prototypical sample for class $c$. Also, $\mathcal{N}(0, \sigma^2)$ is the Gaussian noise added to the neuron weight to generate the synthetic data. $\sigma$ is determined by the noise_level parameter, introducing variability to the synthetic samples to mimic natural data distribution:
$\sigma = $ noise_level.

The training phase (3) involves adapting neuron weights to minimize the distance from the input vectors while maintaining the structure of the input space defined by class labels. The generation phase (7) then uses these adapted neuron weights as centers to generate new data points by adding controlled Gaussian noise, effectively creating synthetic examples that are statistically similar to the original samples but include slight variations to enhance robustness and data privacy. This two-phase approach allows SNG to not only classify and categorize data effectively but also to generate new samples that can be used to augment the original dataset, addressing issues such as data scarcity and enhancing model training without compromising data privacy. Figure 4 depicts the flowchart of the proposed method.

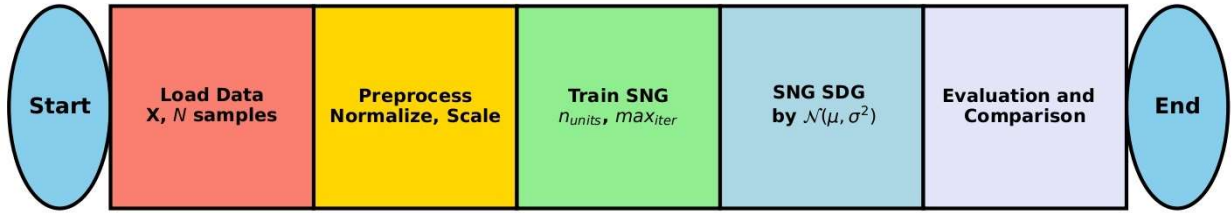

Figure 4. Proposed method's flowchart

## 5. Evaluations and Results
- **Dataset**

We employed two physiological datasets in our experiments for validation as follows. First, we used a standard IEEE emotion recognition dataset called "BRAINWAVE EEG DATASET" [40, 41], which is available online at[4] [42]. This dataset consists of brainwave EEG signals from eight subjects collected in a lab-controlled environment under a specific visualization experiment. The data include simple timestamps followed by the five bands of brainwave signals reading from the five electrodes of the emotive insight sensor: Theta, Alpha, Low Beta, High Beta, and Gamma. More than 10,000 brainwaves were collected. However, after applying several data filtering techniques, including the removal of noise signals and margins from the start and end of each picture showing time, only 1550 brainwaves remained. insider threats by analyzing brain activity through EEG signals involves understanding how certain brain activity patterns will correlate with deceptive or malicious intent, possibly identifying individuals who will pose an insider threat before any malicious actions occur. The concept involves using EEG to detect subconscious or conscious signs of malicious intent or deception in individuals. By analyzing brainwaves, researchers aim to find biomarkers or patterns that indicate a risk of insider threats. Each picture is attached with two main values: Valence, which shows the degree of positive or negative effect the image evokes, and Arousal, which shows the intensity of the effect the image evokes. Images with a valence value equal to one are labeled as zero: ''High Risk''. Images with valence values equal to two and three are labeled as one: ''Medium Risk''. Images with valence values equal to four and five are labeled as two: ''Normal''. Images with valence values equal to six and seven are labeled as three: ''Low Risk''. All images selected as part of this experiment had arousal values of more than five to ensure their intense impact on the participants.

---
[4] https://ieee-dataport.org/documents/brainwave-eeg-dataset

The second dataset is called "Emotional Status Determination using Physiological Parameters Data Set" or, in short, ESD [44], available online by[5] [43]. This dataset is created using the Galvanic Skin Response Sensor and Electrocardiogram sensor of MySignals Healthcare Toolkit. MySignals toolkit consists of the Arduino Uno board and different sensor ports. The sensors were connected to the different ports of the hardware kit, which Arduino SDK controlled. MySignals is a development platform for medical devices and e-Health applications. It is a multichannel physiological signal recorder that measures more than 15 different biometric parameters such as pulse, breath rate, oxygen in blood, electrocardiogram signals, blood pressure, muscle electromyography signals, glucose levels, galvanic skin response, lung capacity, snore waves, patient position, airflow and body scale parameters (weight, bone mass, body fat, muscle mass, body water, visceral fat, Basal Metabolic Rate and Body Mass Index). This novel dataset can be applied for training and evaluating deep learning, machine learning, and data analytics models to deal with binary and multi-class stress and emotion classification problems. The dataset consisted of 253 samples of 14 features of GSR and ECG with different statistical properties. Final columns indicates one of four emotional classes of Fear, Angry, Happy, and Sad. Furthermore, the elicitation was done using 17 videos on all participants, and they reported their emotions on forms after watching them.

- **Classifier and Metrics**

For the classification, we selected XGBoost [45, 46] because we found it the most effective of others during the experiment. XGBoost (Extreme Gradient Boosting) is an optimized distributed gradient boosting library designed to be highly efficient, flexible, and portable. It implements machine learning algorithms under the Gradient Boosting framework, providing a scalable, fast, and accurate method for regression, classification, and ranking problems. For evaluation, seven metrics of accuracy, standard deviation (std), precision, recall, F-1 score, train runtime, and Mean Square Error (MSE) between the original and the synthetic samples have been used. Accuracy is a measure of how often a model correctly predicts the outcome, representing the ratio of correct predictions to the total number of predictions. The std quantifies the amount of variation or dispersion within a set of data values, indicating how spread out the data points are from the mean. Precision, often used in classification problems, measures the accuracy of positive predictions, defined as the ratio of true positive results to the total predicted positives. Recall, also known as sensitivity, assesses the model's ability to identify all relevant instances, calculated as the ratio of true positive results to the actual total positives in the data. The F1-score combines precision and recall into a single metric by calculating their harmonic mean, providing a balanced measure of a model's accuracy, particularly useful when dealing with imbalanced datasets [45, 46]. Train runtime is to evaluate the complexity of each algorithm. Finally, the MSE has been used to evaluate the similarity between the original and the synthetic sample. The evaluation is based on three categories: the original or the baseline, the synthetic data, and a mix of them for all metrics.

---

[5] https://ieee-dataport.org/documents/emotional-status-determination-using-physiological-parameters-data-set

- **Comparison Algorithms**

For comparison, we used four main algorithms in the field of SDG: Conditional Variational Auto Encoders (C-VAE) [47], Conditional Generative Adversarial Networks (C-GAN) [48], Single-Step Diffusion Model [51], and Variational Long Short-Term Memory (V-LSTM) [49, 50]. Conditional comes from the model's ability to generate data conditioned on specific input information, such as labels or other related features.

A Conditional Variational Autoencoder (C-VAE) is a type of generative model that extends the basic Variational Autoencoder (VAE) framework by incorporating conditional parameters, enabling the generation of data with specific attributes. In SDG, C-VAEs are particularly useful for creating complex and diverse datasets that adhere to specified conditions or labels. The model achieves this by conditioning both the encoder and decoder on additional input features, allowing it to learn a conditioned distribution of the data. This makes C-VAEs ideal for tasks where control over certain characteristics of the generated data is crucial, such as generating patient data with specific medical attributes or images with designated object types. Conditional Generative Adversarial Networks (C-GANs) adapt the Generative Adversarial Network (GAN) architecture by incorporating label information into both the generator and discriminator, guiding the data generation process to produce data with specific characteristics. In SDG, C-GANs are valued for their ability to generate highly realistic and detailed samples under controlled conditions. The discriminator in a C-GAN learns to verify not only the authenticity of the generated data but also its alignment with the conditional labels, while the generator strives to produce data that passes the discriminator's tests. This dual-drive mechanism enables C-GANs to create precise and diverse synthetic datasets, which are useful in scenarios where fidelity to real-world data distributions is critical, such as in training machine learning models where real data may be scarce or sensitive. The Single-Step Diffusion Model offers a streamlined approach to SDG for tabular data. This model simplifies the traditional, multi-step denoising process seen in conventional diffusion models by condensing it into a single denoising step. It works by adding Gaussian noise to the original data and then using a neural network to recover the clean, noise-free data. This adaptation is particularly suitable for tabular data, where the relationships between variables can be effectively captured and modeled through a single recovery phase. By focusing on a single-step recovery, the model efficiently learns the underlying data distribution, which is crucial for generating high-quality synthetic datasets that maintain the statistical properties of the original data without the computational complexity of traditional diffusion processes. The V-LSTM for SDG is an advanced technique that integrates variational dropout into the LSTM architecture to enhance its effectiveness in generating synthetic sequential data. By applying the same dropout mask at each time step across the hidden units, this method maintains temporal consistency in dropout application, which is crucial for learning dependencies in sequence data. This consistency allows the LSTM to better model the intricate temporal dynamics and reduce overfitting, resulting in more robust generalization. The key advantage of SDG is that Variational Dropout LSTM can generate high-quality, diverse sequences that closely mirror real-world distributions while managing the risk of overfitting to the training data. This makes it particularly useful for applications where the authenticity and variability of synthetic sequences are critical, such as in financial forecasting, healthcare data simulation, and other areas where sequence data is central.

- **Experiment Setup**

2000 synthetic samples were generated for each dataset and algorithm. All experiments were conducted using 70% training and 30% testing over five runs. Also, all algorithms passed through 100 iterations to synthesize samples. Noise level and batch size were also considered 0.1 and 32 for all algorithms, respectively. Specifically for the SNG, the number of neurons is considered 10, as less than this number brings unreliable classification accuracy with high MSE, and higher values bring no significant performance improvement but increase complexity. Also, the number of samples for the Brain Wave EEG dataset is 1550, which in mixed with synthetic will be 3550, and the EDS dataset covers 253 samples, which in mixed with synthetic data will be 2253 samples. The number of samples to be generated for each class is considered to be equal. Figure 5 illustrates the comparison between a sample signal from Brain Wave EEG data in the original and the synthetic form by the SNG algorithm as a line plot. By looking at the data points, a high level of similarity is visible. However, they are not the same as we achieved an MSE of 0.059 by SNG for this dataset. Also, Figure 6 depicts a scatter plot of features three and four from the ESD dataset regarding original and synthetic samples generated by the SNG algorithm. The similarity between the two plots shows that SNG could successfully capture the data points samples relation of the data and generate corresponding synthetic samples with NGN algorithmic structural distribution.

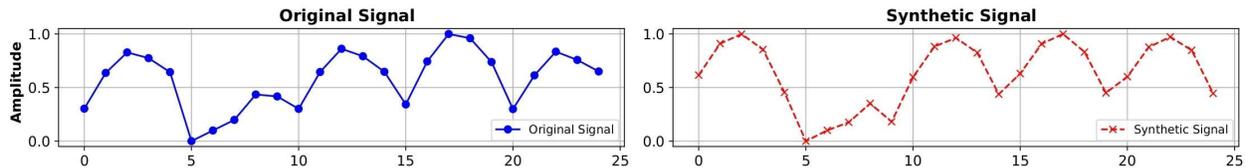

Figure 5. Line plot of a sample from the Brain Wave EEG dataset (left the original and right the synthetic generated by SNG)

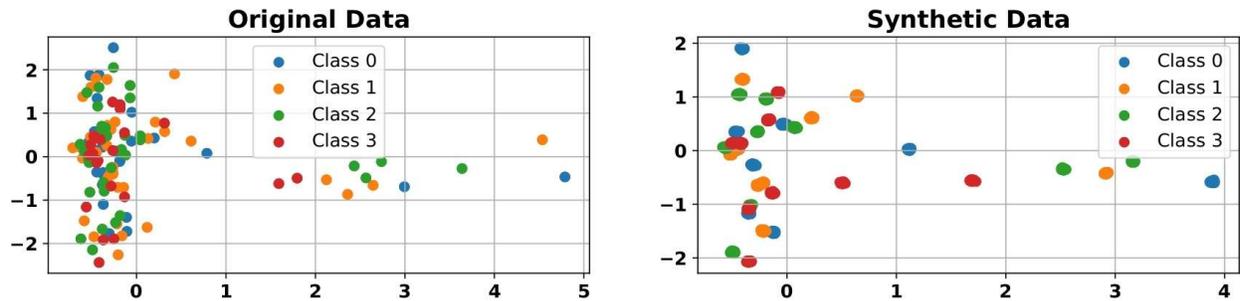

Figure 6. The scatter plot of features three and four from the ESD dataset regarding original and synthetic samples generated by the SNG algorithm (Class 0:Fear, Class 1:Angry, Class 2:Happy, and Class 3:Sad)

Figure 7 represents the t-distributed Stochastic Neighbor Embedding (t-SNE) plot for both synthetic datasets. The t-SNE is a powerful machine learning algorithm used to visualize high-dimensional data by reducing it to two or three dimensions, making it easier to plot and interpret visually [52]. This technique is particularly well-suited for the visualization of datasets with complex structures at multiple scales. t-SNE works by converting similarities between data points to joint probabilities and then minimizing the Kullback-Leibler divergence between the joint probabilities of the low-dimensional embedding and the high-dimensional data. This results in a plot where similar data points are placed close together and dissimilar points are placed far apart, thus revealing intrinsic patterns in the data, such as clusters or groups. The plots reveal several tightly grouped clusters as well as some outliers, suggesting natural groupings within the data. However, the ESD t-SNE plot is more distinctive than the Brain Wave EEG t-SNE plot.

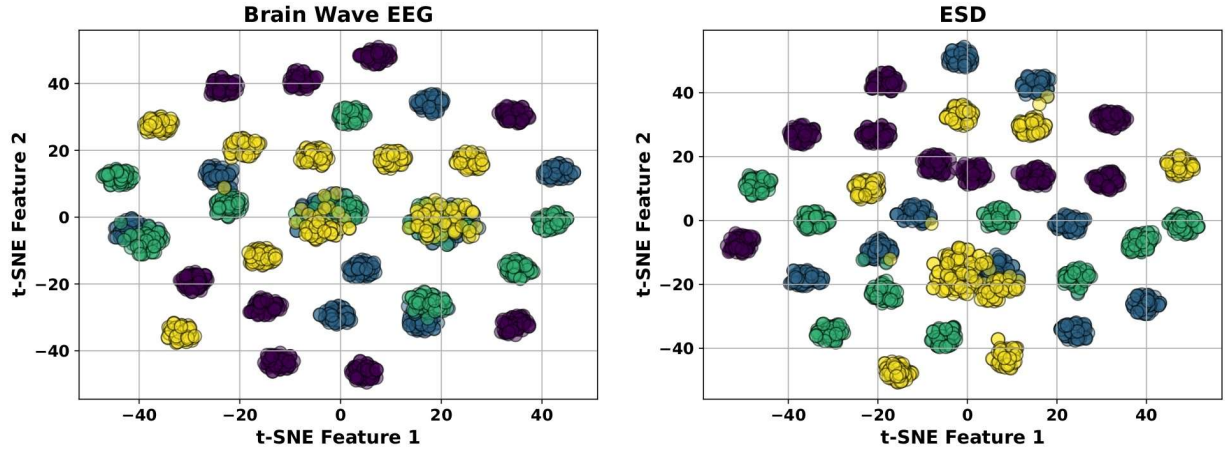

Figure 7. The t-SNE plot of both synthetic datasets (Brain Wave EEG dataset: Class 0: Low Risk, Class 1:Low Medium Risk, Class 2:Medium Risk, Class3: High Risk – ESD dataset: Class 0:Fear, Class 1:Angry, Class 2:Happy, and Class 3:Sad)

Figure 8 depicts the SNG algorithm's loss plot for generating synthetic samples of both datasets. This plot shows the training loss decreasing from around six to below two over 100 iterations. The sharp decline early on suggests that the model quickly learned a significant amount of information from the dataset, and then the rate of decrease slowed, indicating diminishing returns on learning as the training progressed. The training loss for the ESD dataset starts at around three and decreases to around one, also over 100 iterations. Similar to the Brain Wave EEG plot, there is a notable rapid decrease in loss at the beginning, followed by a gradual flattening of the curve, which typically reflects the model reaching a point of stabilization where additional learning provides smaller improvements.

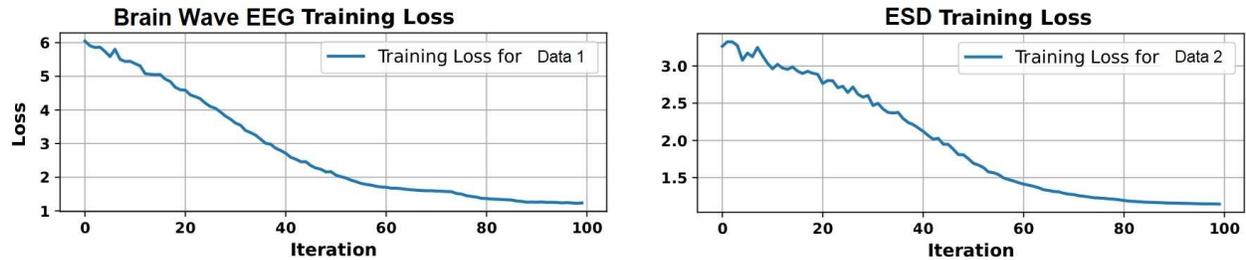

Figure 8. Training Loss Over 100 Iterations for Brain Wave EEG and ESD Datasets

Figure 9 illustrates violin plots for both datasets with different data combinations. A violin plot [53] is a method of plotting numeric data. It is similar to a box plot but with a rotated kernel density plot on each side. This type of plot provides a deeper understanding of the distribution of the data, showing peaks, valleys, and tails more clearly than box plots. The violin shape of the plot displays the density of the data at different values, with the width of the plot representing the frequency of data occurrences at each level. This makes it excellent for comparing multiple distributions, particularly to highlight differences in distribution shape, central tendency, and variability. Violin plots are often used in exploratory data analysis to visualize and compare the distribution of data across different categories or groups. The first row of this figure belongs to the Brain Wave EEG dataset. Plots in this row illustrate that models trained on original data tend to have slightly higher accuracy, maintaining a narrow distribution around 0.95 to 0.98. Synthetic data shows a broader distribution, indicating greater variability in model performance, with accuracies ranging broadly around 0.89 to 0.94. The combined data retains high accuracy similar to the original but with a slightly increased variability compared to the original alone. As for the second row of the ESD dataset, the original data also shows high accuracy but with a wider distribution than seen in the Brain Wave

EEG, suggesting more variability in model performance on this dataset. Synthetic data for ESD also shows high variability but maintains a relatively high accuracy. The combined data seems to perform the best in terms of both median accuracy and consistency, suggesting that combining original and synthetic data may provide a stability benefit in model training for this dataset. Furthermore, Table 2 covers all experiment results for different algorithms using different metrics on both datasets using the XGBoost classifier for the test phase.

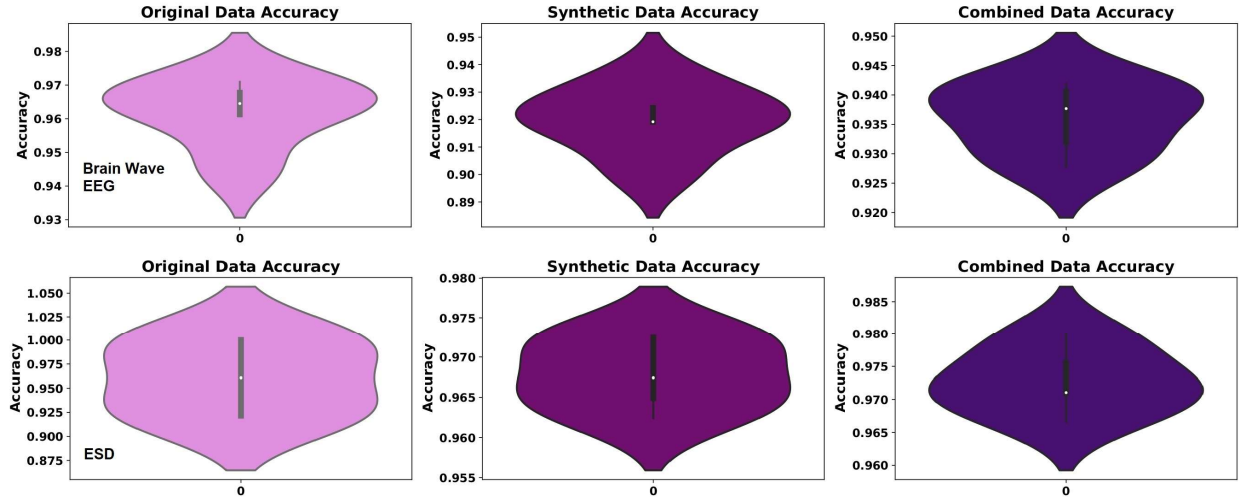

Figure 9. Violin plot of both datasets in different combinations (first row: Brain Wave EEG dataset and the second row ESD dataset)

Table 2. The XGBoost classification results (test phase)

| SDG Method | Metric | Brain Wave EEG Dataset – Baseline: 96.20%, std:0.016 | | ESD Dataset Baseline:95.31 %, std: 0.029 | |
|---|---|---|---|---|---|
| | | Synthetic | Baseline + Synthetic | Synthetic | Baseline + Synthetic |
| C-VAE Converge: itr 73 | Avg Acc | 62.30 % | 78.56 % | 83.85 % | 84.66 % |
| | std | 0.020 | 0.011 | 0.023 | 0.019 |
| | Precision | 0.61 | 0.79 | 0.83 | 0.85 |
| | Recall | 0.62 | 0.79 | 0.84 | 0.85 |
| | F1 Score | 0.62 | 0.79 | 0.84 | 0.85 |
| | Train Runtime | 2 min, 3 sec | - | 35 sec | - |
| | MSE | 0.093 | - | 0.156 | - |
| C-GAN Converge: itr 88 | Avg Acc | 78.10 % | 88.08 % | 90.45 % | 91.57 % |
| | std | 0.095 | 0.062 | 0.025 | 0.018 |
| | Precision | 0.78 | 0.88 | 0.89 | 0.91 |
| | Recall | 0.77 | 0.88 | 0.90 | 0.92 |
| | F1 Score | 0.78 | 0.87 | 0.90 | 0.92 |
| | Train Runtime | 19 sec | - | 17 sec | - |
| | MSE | 0.130 | - | 0.183 | - |
| SS-Diffusion Model Converge: itr 65 | Avg Acc | 86.20 % | 91.06 % | **98.90 %** | **98.33 %** |
| | std | 0.028 | 0.007 | 0.029 | 0.011 |
| | Precision | 0.86 | 0.91 | 0.99 | 0.98 |
| | Recall | 0.85 | 0.90 | 0.98 | 0.98 |
| | F1 Score | 0.86 | 0.90 | 0.98 | 0.98 |
| | Train Runtime | 18 sec | - | 14 sec | - |
| | MSE | 0.186 | - | 0.127 | - |
| V-LSTM Converge: itr 90 | Avg Acc | 50.67 % | 55.01 % | 80.33 % | 82.05 % |
| | std | 0.028 | 0.064 | 0.403 | 0.055 |
| | Precision | 0.50 | 0.54 | 0.80 | 0.81 |
| | Recall | 0.50 | 0.55 | 0.81 | 0.82 |
| | F1 Score | 0.50 | 0.55 | 0.80 | 0.82 |
| | Train Runtime | 50 sec | - | 23 sec | - |
| | MSE | 0.287 | - | 0.301 | - |

| SNG | Avg Acc | **91.90 %** | **93.35 %** | 96.01 % | 97.37 % |
| Converge: itr 81 | std | 0.006 | 0.011 | 0.009 | 0.006 |
| | Precision | 0.91 | 0.93 | 0.95 | 0.96 |
| | Recall | 0.92 | 0.92 | 0.95 | 0.96 |
| | F1 Score | 0.91 | 0.93 | 0.95 | 0.96 |
| | Train Runtime | **8 Sec** | - | **2 sec** | - |
| | MSE | **0.059** | - | **0.101** | - |
| NGN Synthetic Data | Class 1 | [94.6 0.6 0. 0. ] | | [98.8 0. 0.2 0. ] | |
| Confusion Matrix – Class | Class 2 | [ 0. 92.8 4.4 4.4 ] | | [ 0. 91.6 1.8 5.2 ] | |
| names are based on Figure | Class 3 | [ 0. 6. 86.6 6.6 ] | | [ 0.2 2.6 97.6 0.2 ] | |
| seven's caption | Class 4 | [ 0. 4.8 5.6 89.6] | | [ 0. 6.4 1.4 91. ] | |

- **Discussion**

According to the results reported in Table 2, across all SDG methods and both datasets, combining synthetic data with baseline data consistently enhances performance metrics such as accuracy, precision, recall, and F1 score compared to using synthetic data alone. This suggests that synthetic data, while beneficial, is most effective when combined with the original/baseline data due to the fact that it has more samples and is diverse for training. C-GAN and SS-Diffusion Models show particularly strong performance enhancements when synthetic data is combined with baseline data, especially for the ESD dataset where accuracy improves significantly (C-GAN: 90.45% to 91.57%, SS-Diffusion: 98.90% to 98.33%). V-LSTM demonstrates lower accuracy overall, particularly with the Brain Wave EEG dataset. However, there's a notable improvement when synthetic data is used in combination, underscoring the potential for synthetic data to boost weaker models. SNG exhibits high stability and effectiveness, with minimal standard deviation and high accuracy, making it a robust choice across datasets. The standard deviation in performance metrics across methods varies, with some methods, like SNG, showing remarkable consistency. Training runtime also varies significantly across methods, with some, like SNG, being exceptionally quick, suggesting efficiency in training. The confusion matrix for SNG synthetic data indicates high class-specific accuracy, particularly in classes 1 and 4. Misclassifications are mostly contained within adjacent classes, which could point to areas where the model's discrimination between similar classes could be improved. The C-VAE method exhibits moderate MSE values, indicating a fair approximation to the original data but with potential for further refinement. C-GAN, with higher MSE values, shows greater deviation, suggesting that while it can generate diverse data, it may not always closely mirror the original dataset. The SS-Diffusion Model stands out with lower MSE values, indicating that it closely replicates the original data and effectively captures the underlying patterns. V-LSTM shows the highest MSE, indicating significant discrepancies and the least fidelity to the original data among the methods tested. Lastly, the NGN demonstrates low MSE values, suggesting high accuracy and a close match to the original dataset, making it one of the more reliable methods in terms of data generation fidelity. Additionally, it has to be mentioned that 100 iterations were sufficient for all algorithms as both datasets were small-sized, and after 80 iterations, all of them converged.

## 6. Conclusion

This study has significantly advanced the field of emotion recognition using physiological signals by employing an SNG network for synthetic data generation. Our research clearly demonstrates that SNG, while a novel application in this domain, effectively addresses the critical challenge of data scarcity that hampers the development of robust emotion recognition systems. By generating synthetic data that closely mirrors the complex relationships and distributions of real physiological signals, SNG has shown its potential to enhance the accuracy, diversity, and speed of training emotion recognition models. Our experiments indicate that the integration of synthetic data with baseline data consistently improves model performance across various metrics, including accuracy, precision, recall, and F1 score. Notably, SNG exhibited superior stability and minimal variability in performance, making it a robust choice for synthetic data generation across different datasets. This is particularly evident in its high class-specific accuracy and the efficiency of its training process, as reflected in the markedly low mean squared errors and short training times compared to other tested methods such as C-VAE, C-GAN, SS-Diffusion Model, and V-LSTM. The findings affirm that SNG not only successfully generates data that enhances model training but also contributes to more accurate and reliable emotion recognition systems. As we move forward, the application of SNG in other domains of affective computing and beyond holds promising potential to overcome similar challenges of data limitation. This pioneering use of SNG in emotion recognition sets a precedent for further research and development in the field, potentially revolutionizing how synthetic data generation is approached in enhancing human-computer interaction. As for future works, using SNG SDG for other modalities, such as image and body motion, is in progress. Furthermore, integrating SNG with other machine learning models and ensemble methods such as C-GAN, C-VAE, and transformers could enhance the robustness and accuracy of systems designed for emotion recognition data synthesis.